\documentclass{article} % For LaTeX2e
\usepackage{iclr2020_conference,times}
\usepackage{graphicx}
\usepackage{booktabs,multirow} % for professional tables
\usepackage{amsmath}
\usepackage{makecell}
\usepackage{wrapfig}
\usepackage{enumitem}

\usepackage{amsmath,amsfonts,bm}

\def\eqref#1{equation~\ref{#1}}
\def\1{\bm{1}}

\DeclareMathAlphabet{\mathsfit}{\encodingdefault}{\sfdefault}{m}{sl}
\SetMathAlphabet{\mathsfit}{bold}{\encodingdefault}{\sfdefault}{bx}{n}

\usepackage{hyperref}
\usepackage{url}

\title{Amortized learning of neural causal representations}

\author{Nan Rosemary Ke\textsuperscript{*1$\ddagger$},  Jane. X. Wang\textsuperscript{2},
Jovana Mitrovic\textsuperscript{2}, 
Martin Szummer\textsuperscript{2}, Danilo J. Rezende\textsuperscript{*2} }

\newcommand\nnfootnote[1]{%
  \begin{NoHyper}
  \renewcommand\thefootnote{}\footnote{#1}%
  \addtocounter{footnote}{-1}%
  \end{NoHyper}
}

\iclrfinalcopy % Uncomment for camera-ready version, but NOT for submission.
\begin{document}

\maketitle
\begin{abstract}
Causal models can compactly and efficiently encode the data-generating process under all interventions and hence may generalize better under changes in distribution. These models are often represented as Bayesian networks and learning them scales poorly with the number of variables. Moreover, these approaches cannot leverage previously learned knowledge to help with learning new causal models. In order to tackle these challenges, we represent a novel algorithm called \textit{causal relational networks} (CRN) for learning causal models using neural networks. The CRN represent causal models using  continuous representations and hence could scale much better with the number of variables. These models also take in previously learned information to facilitate learning of new causal models. Finally, we  propose a decoding-based metric to evaluate causal models with continuous representations. We test our method on synthetic data achieving high accuracy and quick adaptation to previously unseen causal models.

\end{abstract}

\section{Introduction}
\nnfootnote{\textsuperscript{1}Mila, Polytechnique Montreal, \textsuperscript{2}Deepmind,  \textbf{\textsuperscript{*}Authors contributed equally}, \textsuperscript{$^\ddagger$}Work done during an internship at Deepmind, \texttt{rosemary.nan.ke@gmail.com}}
Neural networks have achieved astonishing success in many machine learning areas by relying on the assumption of independent and identically distributed (iid) data in training and testing settings. In particular, these methods rely on learning correlations from the data and cannot per se identify causal links \citep{de2019causal,goyal2019recurrent,scholkopf2019causality}.  In many settings of interest, e.g.\ domain adaptation, transfer learning and multitask learning, this is a problem as the iid assumption does not hold. On the other hand, causal models efficiently and compactly encode the data-generating process under interventions. 

Previous work in causal learning often focuses on learning representations for causal models in terms of Bayesian networks, e.g.~\citep{heckerman1995learning,cooper1999causal,spirtes2000causation,chickering2002optimal,tsamardinos2006max,shimizu2006linear,pearl2009causality,hoyer2009nonlinear,Peters2011b,hauser2012characterization,goudet2017causal}. While such approaches provide an easily interpretable output, they rely on iterating over all possible graph structures and thus scale poorly with the number of variables. % the number of possible graphs scales super exponentially with the number of variables. 
In addition, it is challenging for Bayesian networks to utilize previously learned information to more efficiently learn new causal graphs in settings such as meta-learning.  Hence, we propose a paradigm shift away from learning Bayesian networks to learning continuous representations of causal graphs using deep neural networks, such that the learned model not only scales better with the number of variables, but can also take into account previously learned information to help learn new causal graphs more efficiently. 

While most research in causal induction focuses on learning a single causal graph at a time, we explore a meta-learning setting, where the algorithm observes many distinct causal graphs during training and is asked to quickly recover a new causal graph during test time.

\begin{figure}[tbh]
\vspace{-1\baselineskip}
\begin{center}
    \includegraphics[width=0.8\textwidth]{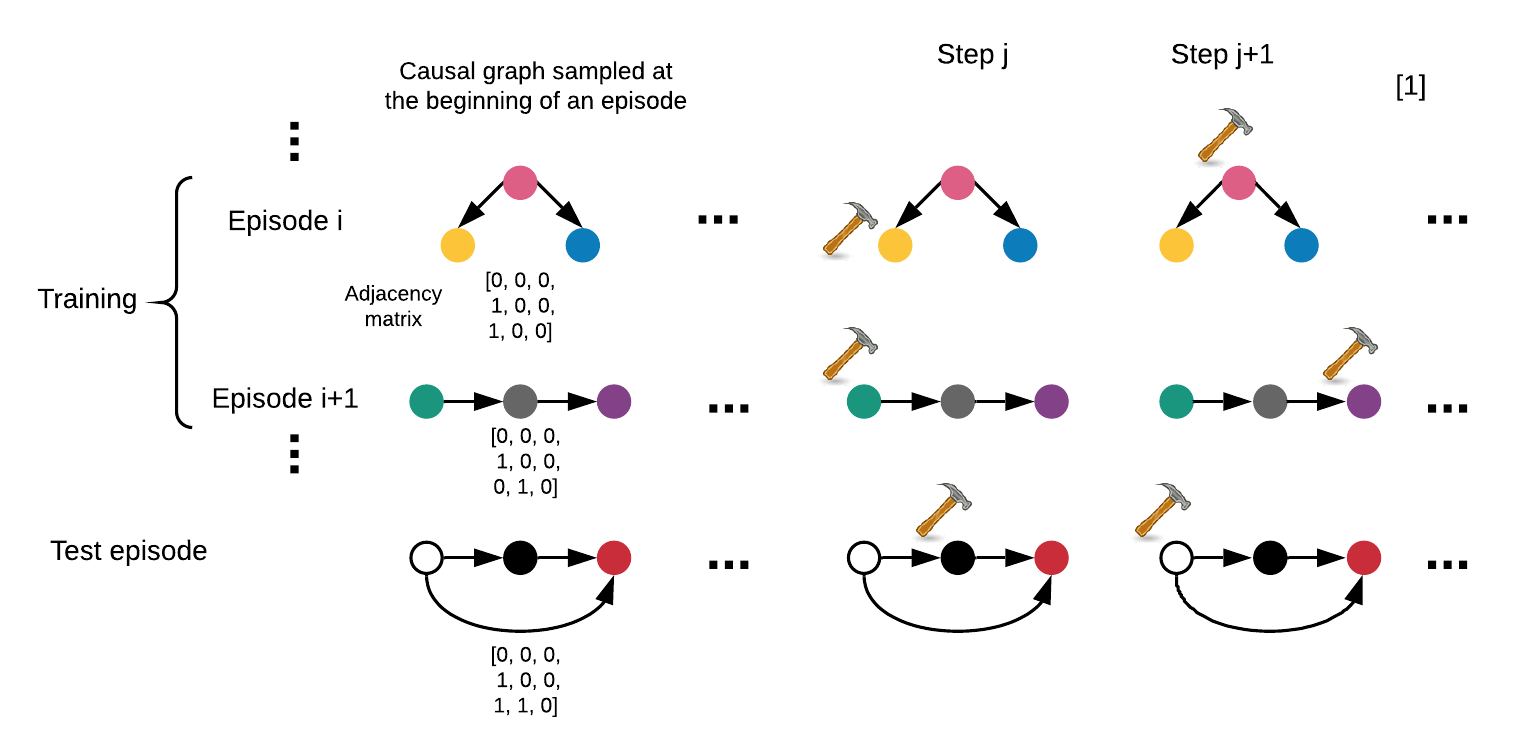}
        \vspace{-1.5\baselineskip}
    \caption{\small Meta-learning task setup, showing an example with $k$ training episodes and $m$ test episodes. At the beginning of episode $i$ (for $i<k$), a new causal graph $G_i$ that governs the data generating process for that episode is sampled. Both the structure and the conditionals of the causal graph are resampled per episode. The structure of the graph can be represented using an adjacency matrix. This is followed by $k$ random and independent interventions (represented by the hammer) on the causal graph $G_i$. The model observes one sample per intervention and $k$ samples in total for an episode, and then evaluated on unseen causal graphs at test time.}
    \label{fig:task_setup}
    \vspace{-1\baselineskip}
        \end{center}
\end{figure}

Motivated by this scenario, we examine a meta-learning setup where learning is composed of  episodes, shown in Figure \ref{fig:task_setup}. At the beginning of each episode, a new causal graph is sampled and the graph governs the data generation during that episode.  At every step $t$ within an episode, a random and independent intervention is conducted on the causal graph and our algorithm receives the node values of a sample under intervention. Note that we assume that there is no time-delay between the intervention and its effect, such that the effect is observed immediately. Thus, during an episode of length $k$ our algorithm observes $k$ samples from the underlying causal graph and each under a different interventions. To facilitate learning, we ask our model to predict outcomes of interventions at every step. Note that this is different from the meta-learning setups in \citet{bengio2019meta,ke2019learning}. In their setting, there is a single causal graph across different episodes and in our setting, there is a distinct causal graph for each episode and hence our algorithm observes many different causal graphs. 
In order to enable learning of causal graphs in this setup, we propose a novel neural network architecture that efficiently learns to encode causal graphs in a compact, continuous representation. 
When learning causal graphs explicitly as discrete structures, it is straightforward to evaluate accuracy by counting how many of the inferred edges match the ground truth.
Our implicit, continuous representation of a causal graph cannot be directly be compared against the ground truth.  One possibility is to compare the predictions of node values under all possible interventions, rather than comparing the edge structure. 
Unfortunately, comparing node predictions scales poorly and would be prohibitively expensive even for medium-sized graphs.  Therefore, we instead learn a decoder from the continuous representation into a graph, and then measure its edge accuracy. Our key contributions can be summarized as follows:
\vspace{-2.5mm}
 \begin{itemize}
     \item We introduce a novel framework for learning to learn causal discovery models using continuous representations via unsupervised losses.
     \item This novel framework that can take in previously learned information to facilitate learning of new causal graphs.
     \item We introduce a decoding metric to evaluate causal models with continuous representations.
 \end{itemize}
\vspace{-4mm}
\section{Related work}
\vspace{-2mm}
Most approaches to causal induction are focuses on learning a single causal graph and can not  easily adapt to the meta-learning setting. For a comprehensive review on most of these methods, please refer to \citet{heinze2018causal}.  We will focus our discussions on causal induction methods that explores a meta-learning setting.

The works of \citet{bengio2019meta} and \citet{ke2019learning} explore a meta-learning setting for causal induction from interventional data. However, their setting is different in that there is a single underlying causal model across the different episodes for both training and test. In our setting, there is a distinct causal model per episode and hence the algorithm observes many different causal models during training and the model is asked to recover unseen causal models during test. The work of \citet{nair2019causal} uses a meta-learning setup similar to ours, however, their model is provided a supervised signal from the groundtruth graph during training, whereas ours can infer the causal structure from an unsupervised learning signal.  Most similar to our work, \citep{dasgupta2019causal} uses meta-learning to learn to make predictions under interventions that suggest (but do not prove) that some causal reasoning is occurring.  However, their approach does not induce a causal graph, neither explicitly nor via a decoding, thus it cannot be used for general causal discovery, but rather is more focused on the reinforcement learning setting.
\vspace{-3mm}
\section{Causal relational networks}
We propose a general framework for learning causal models from interventional data by leveraging the power of neural networks. Unlike most previous methods that learn Bayesian network structure, we employ continuous representations to represent causal graphs. Furthermore, we train our model on many distinct causal graphs which allows it to learn new causal graphs very quickly. In order to evaluate the effectiveness of our method, we measure the ability of our model to predict the outcome of interventions as well as decode the structure of the underlying causal Bayesian network. 
\vspace{-2mm}
\subsection{Problem formulation}\label{problem}
We propose a meta-learning setup to learn causal graphs. As shown in Figure \ref{fig:task_setup}, training occurs in episodes, with a  distinct ground-truth causal graph that governs the data generation for the duration of each episode. Episodes have a fixed length $k$. 
At the beginning of each episode, we first draw a new causal graph $G$, followed by generating one sample for a random intervention for a total of $k$ samples per episode. The resulting intervention samples $x$ along with the index of the intervened variable are fed into the model sequentially. Note that there is no time-delay between the intervention and its effects, such that all effects are observed immediately. The model is asked to predict the intervention outcome and hence learning the causal relationships between variables. For a detailed description, please refer to Section \ref{model_description}.

To evaluate the performance of our model, we cannot directly compare the inferred graph structure to the ground truth, as is usually done when learning Bayesian networks. We have continuous representations that implicitly encode the underlying causal graph. Given that evaluating the predictions under all possible interventions does not scale to large graphs and state spaces, we 
propose to use a decoding-based evaluation metric for continuous representations of causal graphs, similar to~\citep{gregor2019shaping}. Specifically, we propose to evaluate if the discrete graph structure can be decoded from the continuous representation by training a neural network decoder. Thus, at every time step $t$, we feed the hidden state of our model into a decoder trained to output the right causal graph. 
Note that we do not pass gradients from the ground-truth supervision signal back into the decoder.
\subsection{Model Description}\label{model_description}
\begin{wrapfigure}{R}{0.5\textwidth}
    \vspace{-3.5\baselineskip}
    \includegraphics[width=0.5\textwidth]{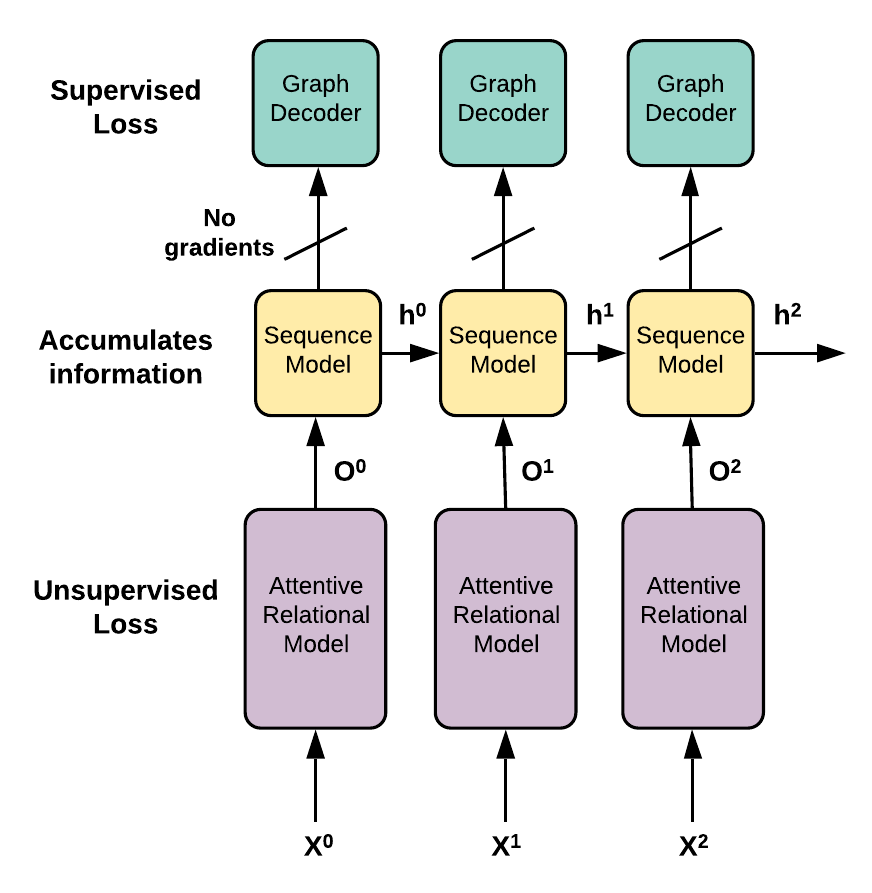}
    \caption{\small Causal relational model with all components: \textit{attentive relational model} that models the data and serves to preserve the relationship between variables; the \textit{sequence model}  accumulates information through time and the \textit{graph decoder}  decodes the graph from the sequence model.}
    \label{fig:model_figure}
    \vspace{-2.5\baselineskip}
\end{wrapfigure}

Our tasks contains sequence data and hence a natural choice of model would be sequence models such as recurrent neural networks (RNNs). However, after some experimentation, we found that it is difficult for vanilla RNN models such as LSTM \citep{hochreiter1997long} to model causal relationships, shown in Section \ref{baseline}. Hence, we propose a new architecture named \textit{causal relational networks} that can better model causal relationships. The model comprises three modules shown in Figure~\ref{fig:model_figure}. 
\begin{enumerate}
    \item The \textit{attentive relational model} models the observed data through a latent representation and serves to preserve the relationship between variables.
    \item The \textit{sequence model} computes the belief state at every  step $t$, such that the belief state at $t$ models the available information up to $t$.
    \item The \textit{graph decoder} outputs the estimated adjacency matrix of the underlying causal graph.
\end{enumerate}
As in a meta-learning setup, training is divided into episodes. Each episode begins with sampling a new causal graph. Data from each episode is sampled from the same causal graph. The samples are read as a sequence as shown in Figure~\ref{fig:model_figure}. Note that the data is not time-series data, since there is no time-delay between interventions and effects from it.
For a causal graph of size $n$, at each time step, one sample vector of size $n$ along  with the intervention information is passed into the model. 
\vspace{-2mm}
\paragraph{Attentive Relational Model.}
A natural choice for encoding the input would be to use a monolithic fully connected encoder with a sequence model (such as a LSTM). However, in our  experiments (in section \ref{baseline}), we  observed such monolithic model does not manage to learn the causal relationships between variables. To tackle this, we propose a novel relational network architecture  inspired by the relational networks used in  \citep{ke2019learning}. This architecture is trained with unsupervised learning. In particular, the model can more appropriately models causal relationships between variables because it is trained to predict, given a particular variable in the sample, the value of all other variables. \begin{wrapfigure}{R}{0.45\textwidth}
    \vspace{-1\baselineskip}
    \includegraphics[width=0.45\textwidth]{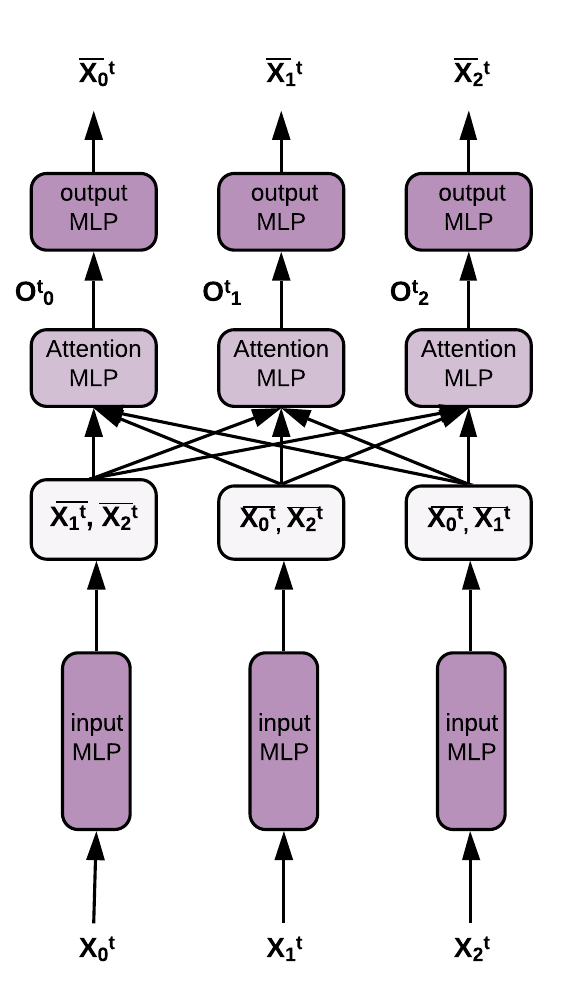}
    \caption{The attentive relational model for inputs with 3 variables. The 3 distinct input MLP takes in each variable $x^t_i$ of the inputs separately and outputs all other variables $x_k^t$, where $k\ne i$. The variables that are direct parents or child of the predicted variable would be able to better predict the variable. Hence, the attention MLP computes a weight based on how well $x_i$ can predict $x_k$ and summarize them into the immediate output $O^t$, which is then passed to the sequence model. The attentive relational model receives an unsupervised loss based on its ability to predict each variable. }
    \label{fig:encoder_model_figure}
    \vspace{-2\baselineskip}
\end{wrapfigure}

Within an episode, let $x^t_{0},x^t_{1},...,x^t_{n}$ denote a particular sample from the causal graph under an intervention at time $t$, $x^t_0$ denote the index of the intervened variable and  $x^t_{i}$ denote  value of the $i$-th variable in the $t$-th sample. For every variable, we instantiate a feedforward neural network 
% (\textit{attention MLP}) 
that takes the values of this variable as input. Note that there are $n$ MLPs in the model that are identical in shape but do not share parameters (see Figure~\ref{fig:encoder_model_figure}). The task of each MLP is to predict the value of the remaining variables (i.e.\  output size $n-1$).  Specifically, let $f_k$ be the MLP that takes as input $x^t_{i}$ and attempts to predict all $x^t_k$, such that $k\ne i$, i.e.\  $x_0, x_1, ...x_{k, k \ne i} = f_k(x_i)$ and hence, there are $n-1$ predictions of each variable $x_i^t$. 

We we augment our architecture with an attention model, based on the intuition that a variable can better predict its direct parents or children (in comparison with  unconnected or distantly connected variables). The attention model can learn how well variable $i$ can predict variable $j$. Thus, for a given variable $x^t_i$, the MLP returns $n-1$ predictions (for all $x_{k, k\ne i}$), which are fed into the \textit{attention MLP} (see Figure~\ref{fig:encoder_model_figure}). This then outputs attention weight $w^t_{i,k}$, with $k$ denoting the prediction of variable $X_i^t$ from $f_k$. Next, we use the attention weights to compute the intermediate prediction $O_i^t$, where this prediction is being passed through a MLP, where it outputs the final predictions of the variable $X_i^t$.  Taking all of this together, the MLPs together with the attention weight encode the relationship between a single variable and the rest of the causal graph. The output for the autoencoder predicts the sample from the intervention distribution $\bar{X^t_0}, ...,\bar{ X^t_{n-1}}$. Note that the model ignores the prediction for the variable that has been intervened upon, since that variable doesn't give us information about the underlying causal graph.
\paragraph{Sequence model.} We use a simple sequence model to accumulate information produced by the encoder and update the belief state of the model. In fact, in our setup, we use a simple cumulative summation operator with no learned parameter,  the sequence model simply adds current information to already accumulated information. The sequence model at time $t$ takes the immediate output $O^t$ from the attentive relational model and add it to the hidden state of the sequence model.
Then the hidden state $h^t$ of the sequence model at time $t$ is updated as $h^{t} = h^{t-1} + O^t$. This simple additive update guarantees that the final representation $h^t$ is invariant with respect to permutations in  the order of presentation of the data points. 
A more complex recurrent model, such as LSTM (\citep{hochreiter1997long}), could also be used but it is not trivial to preserve permutation invariance in this case.

\paragraph{Graph decoder.} Having computed the hidden state $h^t$ of the sequence model, we use a LSTM \citep{hochreiter1997long} to decode the structure of the causal graph. The structure of the causal graph is presented as a  flattened adjacency matrix (a sequence),  and the graph decoder is trained to predict the adjacency matrix conditioned on hidden state $h_t$ from the sequence model. The model is trained using teacher forcing during training and free-running for evaluation. We also ensure that no gradient with supervision signal are passed back into the sequence model.
\section{Experimental setup and results}\label{experiments}
Our experiments aim to evaluate our proposed method for causal discovery in a meta-learning setting. We study  our proposed method for causal induction on synthetic data and we perform experiments to answer the following questions:
\begin{enumerate}[label=\textbf{Q\arabic*}.]
    \item How well can our proposed method recover causal structures?
    \item Can our model leverage previously learned information to more efficiently recover causal structures?
    \item Does the structured relational autoencoder help to extract causal relationships?
    \item Does the accuracy of the decoder reflect the model's ability to recover causal structures?
    \item How much would supervision signal improve the results?
\end{enumerate}
\subsection{Synthetic Dataset}\label{binary_bayesnet}
We explore the setting of binary Bayesian networks with all variables observed. We focus on this setting as a proof of concept for our methodology; extensions to categorical Bayesian networks are straightforward. Intuitively, the causal graphs can be thought of as a soft random Boolean network that is a combination of AND and OR gates, which is motivated by discrete variables combined by noisy AND and OR gates. More formally,  the ground-truth model defined a joint $p(X=x) = \prod_i p\left(X_{i}=x_{i} \mid x_{<i}, w_i\right)$ is formulated as follows, 
\begin{equation}
p\left(X_{i}=x_{i} \mid x_{<i}, w_i\right)=\operatorname{sigmoid}\left[w_{i} \beta\left(\sum_{j<i} U_{j, i} M_{j, i} x_{j}+b_{i}\right)\right],
\end{equation}

where $w_i \in \{0,1\}$ is an intervention mask sampled randomly ($w_i = 0$ corresponds to an intervention replacing the node distribution by an uniform distribution), $U_{i,j}\in\{-1,1\}$ is the coupling potential, $b_i \in \{ 1 - \sum_{j<i}M_{j i}, \sum_{j<i}M_{j i} - 1 \}$ is an integer potential bias, $M_{i,j}\in\{0,1\}$ is the connectivity mask representing the Bayesian net causal graph, and $\beta=5$ is a concentration parameter. This model simulates a Bayesian network over binary variables with randomized interventions. 
It can be thought of as a soft random Boolean network.
For the ease of sampling, we restrict the ground-truth model to a directed acylic graph (DAG). We have fixed the number of variables in the graph to 5 for the experiments.

\textbf{Identifiability} In our setting, all variables are observed, such that there are no latent confounders. All interventions are random selected and are independent and all interventions are hard interventions. It has been shown in \citep{eberhardt2012number} that the true causal graph is identifiable in principle under our setting.

We perform the same hyper-parameter search for our baselines as well as our model. We use a batch size of 8, learning rate of $0.0002$, all models are trained with the Adam optimizer \citep{kingma2014adam} for $20000$ iterations. For the attentive relational model, we set the (encoder) MLP to have 2 hidden layers of size 128 and 64 with a ReLU \citep{nair2010rectified} in between. The relational encoder final output size is $64$. The graph decoder LSTM has a hidden state size of $128$.

\begin{figure*}
    \centering
    \includegraphics[scale=0.35]{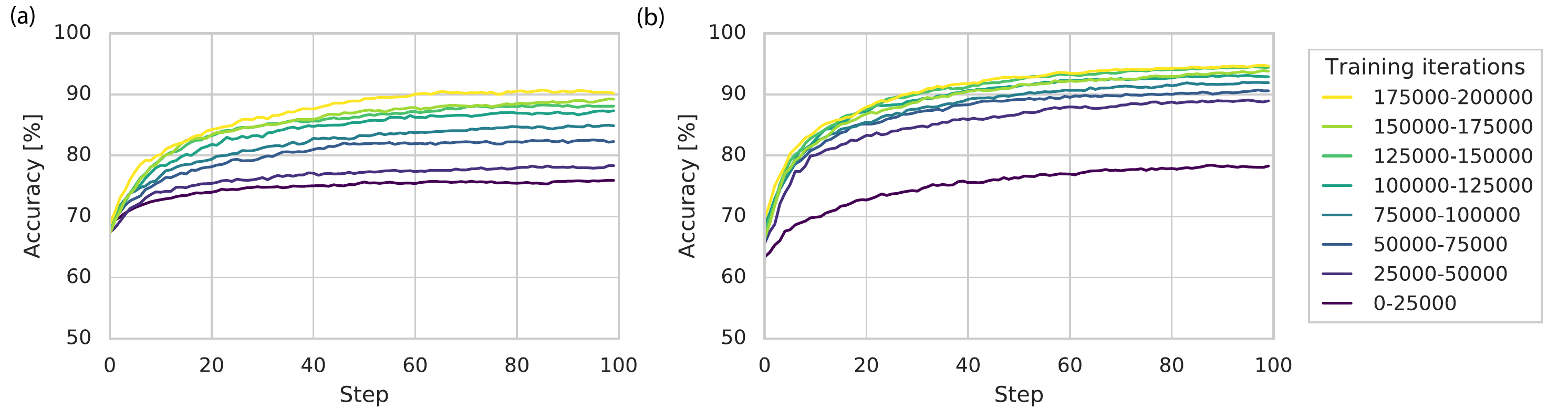}
    \caption{Accuracy for edge probability over the course of the episode, binned by different stages of training, for \textbf{a)} our causal relational network (CRN) model, and \textbf{b)} CRN with supervised graph learning. The X-axis represents step within an episode and the colour-coded curves represent different stages of training.}
    \label{fig:learn_curve}
\end{figure*}

\textbf{Evaluation metrics} We evaluate our model on the accuracy of the decoded graph with respect to the ground-truth graph. The accuracy is defined to be the percentage of decoded edges that align with the ground-truth edge. The edge is said to be aligned with the ground-truth edge if both of the \textit{argmax} of the Bernoulli distribution over the learned edge and the ground-truth edge matches. 
\subsection{Experimental results}\label{our_model}

\textbf{Q1. Recovering causal structure.} As a distinct causal graph is sampled at the beginning of each  episode, the edge accuracy for the models should naturally improve  across the course of an episode, for different stages of training. The model has learned to extract information about the causal graph structure from intervention samples, reflected in the improvements in edge accuracy as the model observes more intervention data from the causal graph.  In the beginning of the episode, the model only knows the underlying causal graph is a DAG. As the model observes more intervention samples from the same graph, it can quickly infer underlying causal graph. As shown in Figure~\ref{fig:learn_curve}a, towards the end of training (yellow curve), our model achieves $90\%$ in edge accuracy just after seeing 100 samples from the new graph. 

\textbf{Q2. Leveraging previously learned information.} Shown in Figure~\ref{fig:learn_curve}a, our model quickly learned to leverage previously learned information to more efficiently learn new causal models. In fact, the model quickly learns that the underlying causal graph is a DAG and  hence achieving over $60\%$ accuracy even at the beginning of each episode.  The model then learns to decode the true causal graph from intervention samples throughout training, this is reflected in the improvement in edge accuracy during the episode throughout the training process.

\subsubsection{Baseline Models}\label{baseline}

There are  3 main components to our model as shown in Figure~\ref{fig:model_figure}. 
First, we evaluate if the \textit{attentive relational model} is necessary to capture causal relationships. Hence, we evaluate against a monolitic baseline model such as a fully-connected MLP, this then forms our LSTM baseline model.  We then evaluate the amount of performance gain the model could achieve by receiving  \textit{supervision signal} from our graph decoder. Lastly, in order to verify that the information for the conditional \textit{graph decoder} for decoding the graph is indeed contained in our model, we compare the performance of our model to one with zero information in its input to the graph decoder. The sequence model that accumulates information across time is fairly straightforward and hence we have not performed comparisons against this component.

\textbf{LSTM baseline.} Our model uses a relatively sophisticated relational network to extract information in the input. In order to verify the necessity of this component, we compare our encoder to a LSTM with naive fully connected feed-forward neural network (MLP). For simplicity, we call this the LSTM baseline. For this baseline, we no longer need an autoencoder to process the input, as the autoencoder could just be an identity model. As there are no autoencoder (with its own decoder), the baseline is missing the unsupervised learning signal. To compensate for this, we pass the gradients from the supervised graph decoder to the MLP encoder. The baseline model has an easier job, since it could learn directly from supervision signal. 
\begin{figure*}
    \centering
    \includegraphics[width=0.7\textwidth]{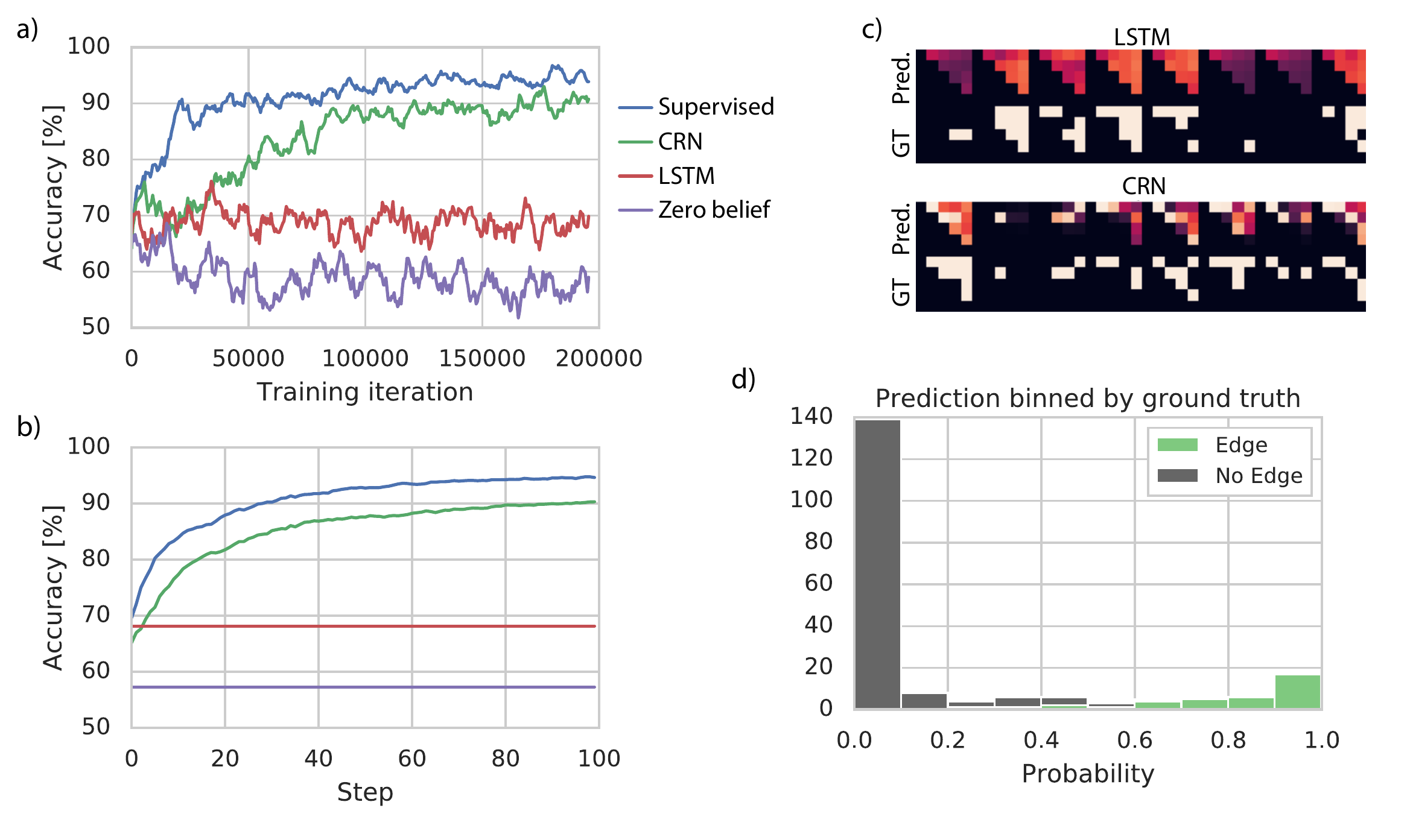}
    \caption{Comparing performance for all models. \textbf{a)} Accuracy of edge prediction after seeing the entire episode of samples, over the course of training. \textbf{b)} Edge accuracy over the course of the episode, averaged over 50 episodes at the end of training. \textbf{c)} Predicted causal graph (adjacency matrix format) vs.\  ground-truth graph for the (upper) LSTM baseline and (lower) CRN. \textbf{d)} Histogram of predicted edge weights at end of episode for ground-truth edges of weight = 1 vs.\  weight = 0, CRN model. CRN = Causal relational network; zero belief = belief state set to 0.}
    \label{fig:compare_results}
\end{figure*}

\textbf{Q3. Is the relational autoencoder necessary?}
Interestingly, as shown in  Figure~\ref{fig:compare_results}, though the baseline model has an easier task, the edge prediction accuracy is only around $70\%$ throughout training as compared to the $90\%$ reached by our model as shown in Figure~\ref{fig:compare_results}. The baseline model neither learns within an episode (as it observes more samples), nor does it learn with more training.

\paragraph{Q4. Does the accuracy of the decoder reflect the model's ability to recover causal structures?} %  Zero belief state baseline.}
Next, we verify that the belief state in our model contains useful information for decoding the structured causal graph. This baseline uses the same relational encoder-decoder and graph decoder from our original model; the  only difference is that the belief state is set to zero. If this baseline could also learn to decode the structured causal graph, it would indicate that it was not the information in the belief state that helped to decode the structured causal graph, it was rather the graph decoder itself.
Figure~\ref{fig:compare_results} shows that the decoder was not in fact able to decode the causal graph structure. The accuracy remains around and below $60\%$ throughout training, compared to the $90\%$ that our model achieved at the end of training (Figure~\ref{fig:learn_curve}a), indicating that the belief state was crucial to decoding the causal graph.

\paragraph{Q5. Supervised graph learning.}\label{supervised_graph}
We next evaluate the amount of performance gain the model gets by receiving supervision signal from the graph decoder. This is achieved by removing the stop gradient from the belief state that is passed in as inputs to the graph decoder. The results are shown in Figure~\ref{fig:learn_curve}b. The model with supervision signal achieved $95\%$ accuracy as compared to the $90\%$ achieved by the unsupervised model.

\section{Conclusion}

In this paper, we presented a new framework for learning and evaluating learned causal models using neural networks. In particular, we have introduced a novel neural architecture \textit{causal relational networks}, and proposed to use a decoding-based metric for evaluation. We have tested our methodology on synthetic data and have shown that it can accurately learn causal structures. Furthermore, we have shown that causal relational networks can very quickly learn new causal models.

\section{Acknowledgements}
The authors would like to thank Theophane Weber, Lars Buesing, Silvia Chiappa, Charles Blundell, Anirudh Goyal, Olexa Bilaniuk and  Alex Lamb for their useful feedback and discussions.

% \clearpage
\bibliography{iclr2020_conference,causality}
\bibliographystyle{iclr2020_conference}

% \appendix
% \section{Appendix}

\end{document}